\pdfoutput=1
\documentclass[10pt,twocolumn,letterpaper]{article}
\usepackage{iccv}
\usepackage{times}
\usepackage{epsfig}
\usepackage{graphicx}
\usepackage{amsmath}
\usepackage{amssymb}
\usepackage{epstopdf}

\iccvfinalcopy 

\def\iccvPaperID{1350} 

\ificcvfinal\fi

\begin{document}

\title{High-Quality Correspondence and Segmentation Estimation \\ for Dual-Lens Smart-Phone Portraits}

\author{Xiaoyong Shen \quad Hongyun Gao \quad Xin Tao \quad Chao Zhou \quad Jiaya Jia\\
The Chinese University of Hong Kong\\
{\tt\small \{xyshen,hygao,xtao,zhouc,leojia\}@cse.cuhk.edu.hk}
 }

\maketitle

\begin{abstract}
Estimating correspondence between two images and extracting the foreground object are two
challenges in computer vision. With dual-lens smart phones, such as iPhone 7Plus and
Huawei P9, coming into the market, two images of slightly different views provide us new
information to unify the two topics. We propose a joint method to tackle them
simultaneously via a joint fully connected conditional random field (CRF) framework. The
regional correspondence is used to handle textureless regions in matching and make our
CRF system computationally efficient. Our method is evaluated over 2,000 new image pairs,
and produces promising results on challenging portrait images.
\end{abstract}

\section{Introduction}

It is convenient now to capture and share photos. It is reported that over one billion
new images \cite{instgram,Shen2016EG} are shared every day over Internet and most of them
are portraits \cite{Dailytech,Telegraph}. With the production of new dual-lens smart
phones, a special way for two-image capturing becomes available for common users, which
actually provides more intriguing information for many photo-related applications.

\begin{figure*}[t]
\centering
\begin{tabular}{@{\hspace{0.0mm}}c@{\hspace{1mm}}c@{\hspace{1mm}}c@{\hspace{1mm}}c@{\hspace{0mm}}}
\includegraphics[width=0.24\linewidth]{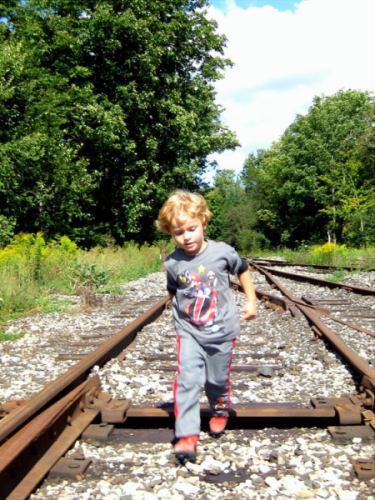}&
\includegraphics[width=0.24\linewidth]{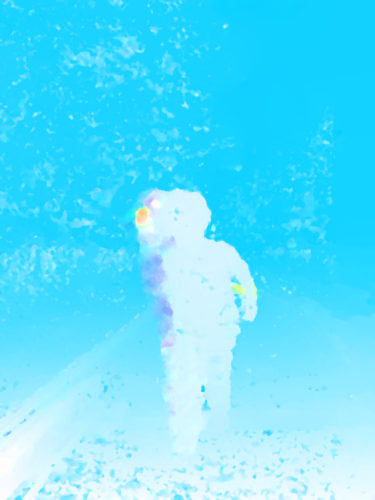}&
\includegraphics[width=0.24\linewidth]{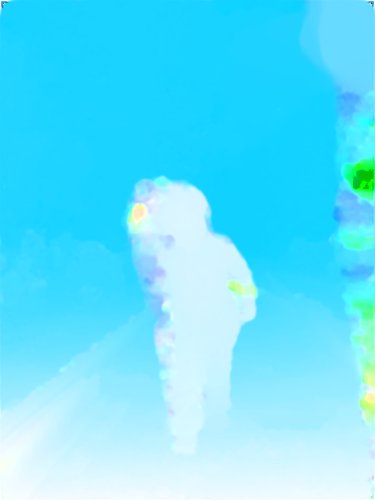}&
\includegraphics[width=0.24\linewidth]{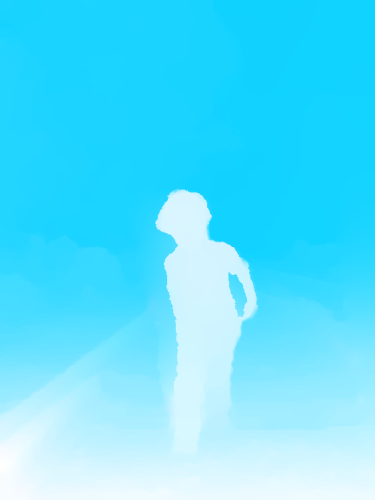}\\
\small(a) Reference & \small (b) MDP Flow & \small(c) LDOF Flow & \small(d) Our Refined\\
\includegraphics[width=0.24\linewidth]{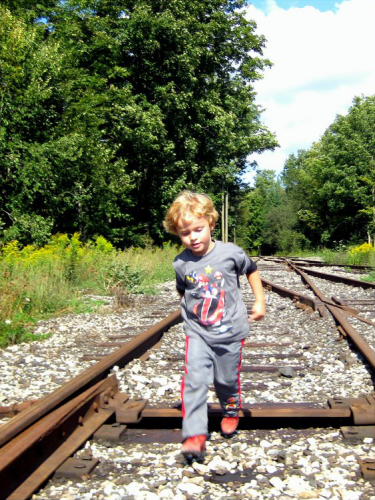}&
\includegraphics[width=0.24\linewidth]{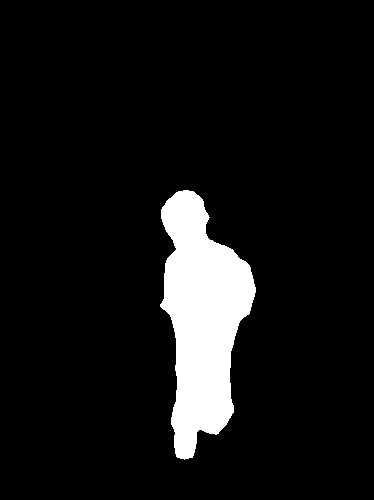}&
\includegraphics[width=0.24\linewidth]{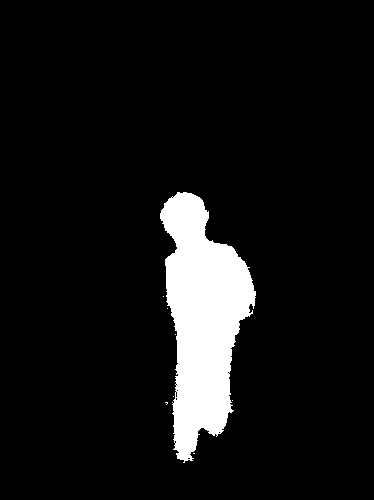}&
\includegraphics[width=0.24\linewidth]{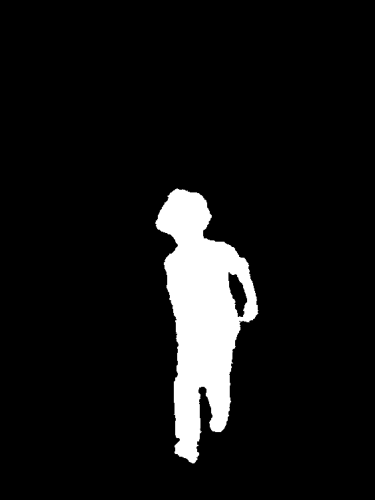}\\
\small(e) Input & \small (f) FCN Segmentation& \small(g) CRFasRNN Segmentation& \small(h) Our Refined\\
\end{tabular}
\caption{Optical flow and semantic segmentation on dual-lens images. (a) and (e) are the
input. (b) and (c) show MDP \cite{XuJM12} and LDOF \cite{BroxLOD10} estimates
respectively. (f) is the FCN \cite{Long2014} segmentation result and (g) is the CRFasRNN
\cite{Zheng2015_crfrnn} result. (d) and (h) are our estimated correspondence and
segmentation respectively. } \label{fig:teaser}
\end{figure*}

It seems a well-studied problem in computer vision that the two-camera output can be used
to estimate depth with pixel correspondence established by optical flow estimation
\cite{HornS81,XuJM12} or stereo matching \cite{Sunstereo2003,Hosni2011}. Meanwhile it is
also known in this community that producing pixel-level-accurate results is still
difficult due primarily to diverse and complex content, textureless regions, noise, blur,
occlusion, etc. An example is shown in Figure \ref{fig:teaser} where (a) and (e) are the
input from a dual-lens camera. (b) and (c) show optical flow estimates of MDP
\cite{XuJM12} and LDOF \cite{BroxLOD10} where errors are clearly noticeable. These types
of errors are actually common when applying low-level image matching.

In this paper, we exploit extra information in dual-lens images to tackle this
challenging problem on portraits. We incorporate high-level human-body clues in pixel
correspondence estimation and propose a joint scheme to simultaneously refine pixel
matching and object segmentation.

\vspace{0.1in}\noindent\textbf{Analysis of Correspondence Estimation~~} Dual-lens images
could be unrectified and with different resolutions. We thus resort to optical flow
estimation instead of stereo matching for correspondence estimation. As briefly discussed
above, several issues influence these methods even with robust outlier rejection schemes
\cite{BroxBPW04,BlackA96,WedelRVBFC08,ZachPB07}. Complicated nonlinear systems or
discrete methods \cite{LempitskyRR08,Chen2013pmflow,BaoYJ14flow} have their respective
optimization and accuracy limitations.

\vspace{0.1in}\noindent\textbf{Difficulty of Semantic Segmentation~~} About semantic
segmentation, state-of-the-art methods are based on fully convolutional networks (FCN)
\cite{Long2014}, which generate an per-pixel prediction score on all classes.
Hierarchical convolution, pooling, rectification and deconvolution layers are adopted in
the network. Even this advanced technique, semantic segmentation is still a challenging
problem in terms of creating very accurate object boundaries. For the example shown in
Figure \ref{fig:teaser}(f), the small background area near the boy's left arm is labeled
as foreground. Although CRFs are applied to incorporate original image structure
\cite{Zheng2015_crfrnn,Chen2014_deeplab}, improvement is limited as shown in (g)
\cite{Zheng2015_crfrnn}. The reason is that the CNNs predicted score is already wrong in
this case.

\vspace{0.1in}\noindent\textbf{Our Approach and Contribution~~} We propose a joint update
method for portrait photos, taking initialization of simple optical flow estimates and
FCN \cite{Long2014} segments. Then we form a joint fully connected conditional random
fields (CRF) model to incorporate mutual information between correspondence and
segmentation features. To make optimization tractable, we propose {\it regional
correspondence} to greatly reduce CRF solution space. As a result, less then $40$ labels
are produced for effective inference. Our method also handles textureless and outlier
regions to improve estimation. To evaluate our approach, we collect 2,000 image pairs
with labeled segmentation and correspondence. Our experiment shows that this method
notably improves the accuracy compared with previous optical flow estimation and semantic
segmentation approaches respectively.

\section{Related Work}

We briefly review optical flow estimation and image segmentation methods. Since both
areas involve large sets of prior work, we only select related methods for discussion.

\vspace{0.1in}\noindent\textbf{Optical Flow Methods~~} For image pairs captured in the
same scene with intensity or gradient constancy, their correspondence can be computed
with the variational model \cite{HornS81}. The involved data terms are used to satisfy
color or gradient constancy \cite{BroxBPW04,BruhnW05,Zimmer09}. Regularization terms can
achieve piece-wise smooth results. The terms are usually formed by robust functions
\cite{BroxBPW04,BlackA96,WedelRVBFC08,ZachPB07}.

Sparse descriptor matching is incorporated in the variational framework to handle large
motion. Representative methods include those of \cite{BroxLOD10} and
\cite{Weinzaepfel2013}. The method of \cite{XuJM12} fuses feature match in each
coarse-to-fine pyramid scale. The variational model is nonlinear, which might be stuck in
local minima when initialization is not appropriate.

Besides the variational model, nearest-neighbor field (NNF) strategies, such as
PatchMatch \cite{BarnesSFG09,BarnesSGF10}, are also applied. Chen \etal
\cite{Chen2013pmflow} estimated a coarse flow by PatchMatch and refined it by model
fitting. To improve PatchMatch quality, Bao \etal \cite{BaoYJ14flow} developed the
edge-preserving patch similarity cost to search for the nearest neighbor. Recently,
multi-scale NNF methods were proposed in \cite{Bailer2015}.

The motion information is also applied to object segmentation as discussed in
\cite{WeissA96,UngerWPB12,SunWSPB13}. However, these methods need many frames to produce
a reasonable result.

\vspace{0.1in}\noindent\textbf{Image Segmentation Approaches~~} Interactive image
segmentation was developed around a decade ago. These methods take user specified segment
seeds for further optimization by graph cuts or CRF inference. Representative methods
include graph-cut \cite{Boykov2001interactive}, Lazy Snapping \cite{Li2004_lazy}, Grabcut
\cite{Rother2004_grabcut}, and paint selection
\cite{Liu2009_paintselection,AdobePhotoshop}.

Recently, deep convolutional neural networks (CNNs) achieve great success in semantic
segmentation. CNNs are applied mainly in two ways. The first is to learn image features
and apply pixel classification \cite{Arbelaez2012,Mostajabi2014,Farabet2013}. The second
line is to adopt an end-to-end trainable CNN model from input images to segmentation
labels with the fully convolutional networks (FCN) \cite{Long2014}.

To improve performance, DeepLab \cite{Chen2014_deeplab} and CRFasRNN
\cite{Zheng2015_crfrnn} employed dense CRF to refine predicted score maps. Liu \etal
\cite{LiuDPN2015} extended the general CRFs to deep parsing networks, which achieve
state-of-the-art accuracy in the VOC semantic segmentation task \cite{Everingham2010}.
Most CNNs are constructed hierarchically by convolution, pooling and rectification. They
aim at challenging semantic segmentation with class labels. In terms of segmentation
quality, interactive segmentation still perform better since users are involved.

\vspace{0.1in}\noindent\textbf{Segmentation and Correspondence~} To further improve
correspondence estimation, methods of \cite{SunSB10,SunSB12,SunWSPB13,Sevilla-LaraSJB16,HurR16} utilized images
layer or segment information. These methods model the correspondence in each layer and
then use the correspondence to infer layer segmentation. A joint model with
correspondence and layer estimation is formed, which is optimized by
Expectation-Maximization (EM). Similar strategies were also employed in stereo matching
\cite{XiongCJ09}. It was found optimization of these models is time consuming and the
segments (or layers) are not that semantically meaningful. Recently, Bai \etal
\cite{BaiLKU16} employed the semantic segmentation to refine the flow field; but no
segment refinement by optical flow is considered.

\section{Motivation of Our Approach}

\begin{figure}[t]
\centering
\includegraphics[width=0.98\linewidth]{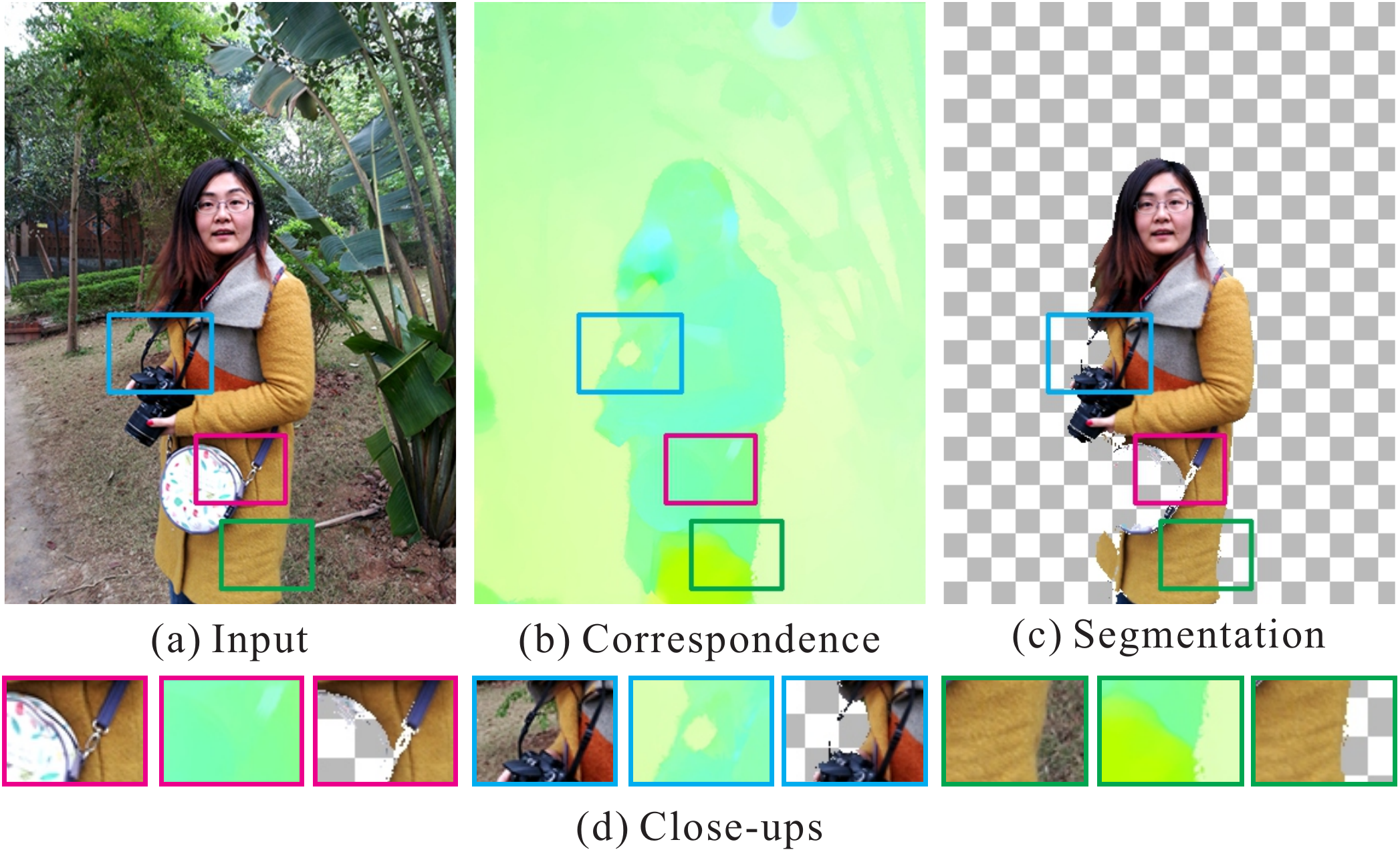}\\
\caption{Discrepancy among input image, correspondence, and segmentation. (a) is the
input image. (b) and (c) show the correspondence and segmentation maps respectively. (d)
gives the close-ups.  } \label{fig:issues}
\end{figure}

Joint update of correspondence and segmentation is difficult because of the domain-level
discrepancy among input image, estimated correspondence, and predicted segmentation. We
show an example in Figure \ref{fig:issues} where (a) is the input image, (b) is the
correspondence result of Horn-Schunck flow method \cite{HornS81} and (c) shows the
segmentation result by FCN \cite{Long2014}. The difference is on the following folds.

\begin{itemize}

\item \textbf{Small Structure~~}
Compared with interactive segmentation, semantic segmentation do not perform accurately
as there exist many small structures in the image. On the contrary, optical flow methods
work better on them. The blue rectangles in Figure \ref{fig:issues}(d) show the
difference.

\item \textbf{Human Belonging and Accessories~~}
Belonging and accessories on human bodies are excluded when performing classification, as
people and other objects are separate into different categories. An example is shown in
red rectangles in Figure \ref{fig:issues}(d) where the bag is excluded. It is not ideal
for portrait images where accessories are part of human bodies.

\item \textbf{Textureless Regions~~}
Correspondence estimation methods may fail in textureless regions. However, segmentation
is less sensitive to them (see green patches in Figure \ref{fig:issues}(d)).

\item \textbf{Complex Background~~} Complex image background incurs extra difficulty for
these methods, which will be detailed later.
\end{itemize}

These discrepancies show that joint refinement is nontrivial for fusion of
different-domain information. Further, the large solution space with continuous
correspondence makes refinement intractable. Our method splits the large solution space
into several regionally accurate parts. With the new form, we achieve the goal via an
efficient fully connected CRF model with a small number of labels.

\begin{figure*}[t]
\centering
\includegraphics[width=0.96\linewidth]{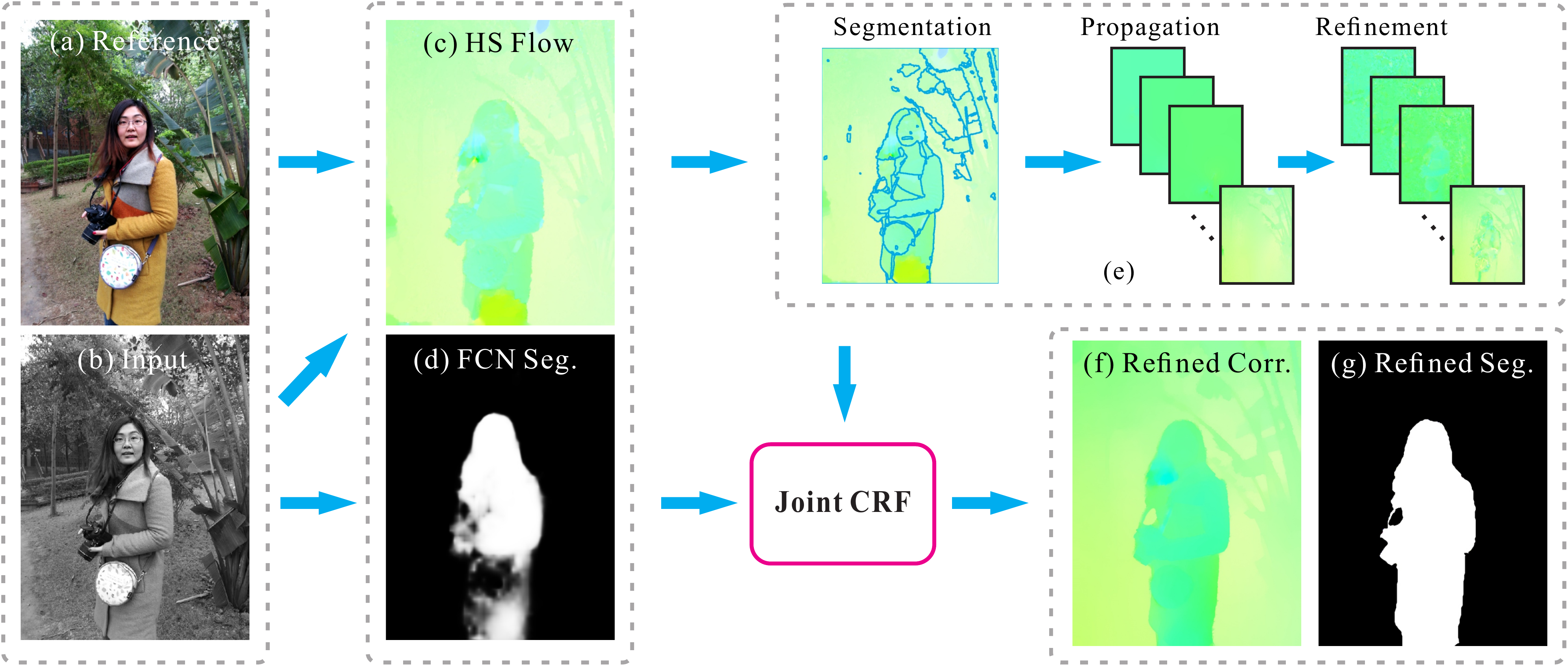}\\
\caption{Illustration of our method. (a) and (b) are the reference and input images
respectively. (c) and (d) are the Horn-Schunck flow \cite{HornS81} and FCN
\cite{Long2014} segmentation results. (e) shows our regional correspondence. (f) and (g)
are the refined results by our joint CRF model.  } \label{fig:illustration}
\end{figure*}

\section{Our Approach}

We estimate pixel correspondence $w$ between images $I_1$ and $I_2$ captured from a
dual-lens smart-phone. Denoting by $p$ the pixel coordinate, displacement $w_p$ is to let
pixel $p$ in $I_1$ correspond to $p+w_p$ in $I_2$. Besides estimating the correspondence
$w$, we also aim for inferring portrait segmentation mask $m$, where $m_p=1$ indicates
the person (i.e., foreground) and $m_p=0$ means background.

We construct a joint CRF model. As illustrated in Figure \ref{fig:illustration}, our
method starts from fast Horn-Schunck flow \cite{HornS81} and FCN segmentation
\cite{Long2014} results. We first estimate regional correspondence for initialization and
then form the joint updating scheme.

\subsection{Regional Correspondence}

Image correspondence is estimated regarding image content. To simplify computation, we
adopt regional correspondence as shown in Figure \ref{fig:illustration}(e). Regional
correspondence is a set of correspondence maps denoted as $\{w^i|i=1,...N\}$ where $N$ is
number of estimated regional correspondence. For each $w^i$, there exist some regions
whose correspondence is accurate. Thus, the final correspondence map $w$ can be computed
by a labeling process considering matching error and correspondence field smoothness.
There are mainly two advantages of the regional correspondence.
\begin{itemize}
\item
Initialization can be set appropriately for each regional correspondence to avoid the
local minimum problem.
\item
Refinement can be achieved by regional correspondence selection to save much computation
time.
\end{itemize}

\vspace{0.1in}\noindent\textbf{Determining Regional Correspondence~~} We compute regional
correspondence by weighted-median-filter-refined \cite{ZhangXJ14wmf} Horn-Schunck flow as
shown in Figure \ref{fig:illustration}(c). The flow field is partitioned into regions
according to motion boundary using the method of \cite{WeinzaepfelRHS15} according to
color and flow features. Regions with similar flow are merged while those completely
different from neighboring regions are discarded as outliers. We apply the very fast
convolutional pyramid \cite{FarbmanFL11convp} to propagate flow to the whole image. The
propagated regional correspondence labels the final result by fusion \cite{XuJM12}.

To improve sub-pixel accuracy, we refine each regional correspondence by the variational
framework \cite{BroxBPW04}. It, in general, can only improve accuracy near edges but not
reliable correspondence for textureless regions, as the data term constraint is not
discriminative enough. We thus only update the regional correspondence in the finest
scale.

\begin{figure}[t]
\centering
\begin{tabular}{@{\hspace{0.0mm}}c@{\hspace{1mm}}c@{\hspace{1mm}}c@{\hspace{1mm}}c@{\hspace{0mm}}}
\includegraphics[width=0.24\linewidth]{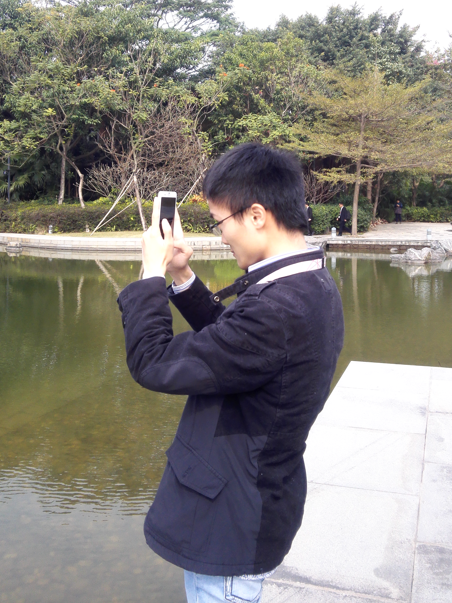}&
\includegraphics[width=0.24\linewidth]{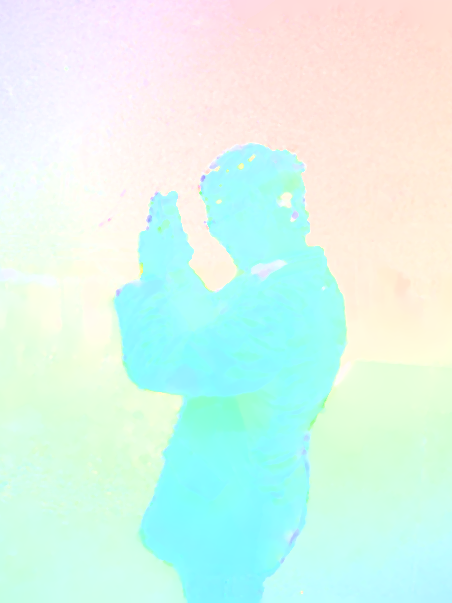}&
\includegraphics[width=0.24\linewidth]{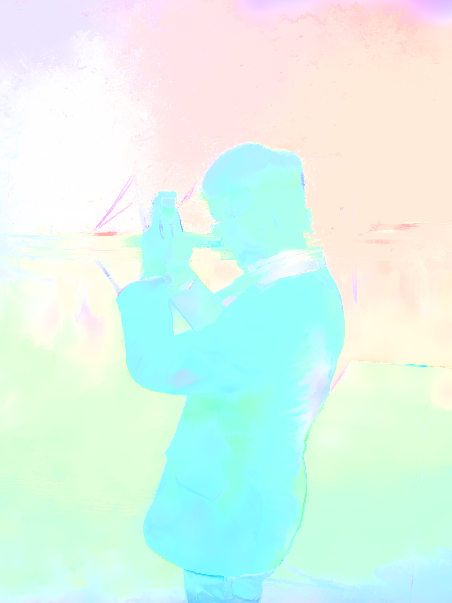}&
\includegraphics[width=0.24\linewidth]{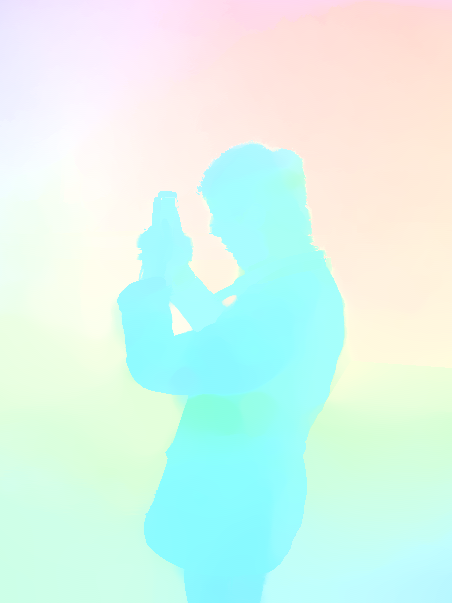}\\
\small(a) Input & \small (b) MDP & \small(c) EPPM & \small(d) Ours\\
\end{tabular}
\caption{Correspondence estimation in textureless regions. (a) shows the input with
textureless regions. (b-c) are MDP \cite{XuJM12} and EPPM \cite{BaoYJ14flow} flow
estimates respectively. (d) is our result. } \label{fig:textureless}
\end{figure}

\vspace{0.1in}\noindent\textbf{Analysis~~} Correspondence propagation is important to
handle textureless regions. We show an example in Figure \ref{fig:textureless}. For the
textureless region between the arms in (a), flow estimation is erroneous as shown in (b).
The PatchMatch-based method \cite{BaoYJ14flow} works better in this region but presents
errors in other area as shown in (c). Our estimate in (d) is from the regional
correspondence by fusion \cite{XuJM12}, which achieves overall better quality. The reason
is that background-propagated regional correspondence gives extra information. In
addition, the number of partitioned regions is small due to sparsity of image content. In
our experiments, $10$ correspondence regions are enough to produce usable results.

\subsection{Joint Refinement Model}
With regional correspondence $w^i$ and FCN predicted initial segmentation as shown in
Figure \ref{fig:illustration}(e) and (d), we adopt a fully connected CRF to improve them.
Our model is formulated as
\begin{align}
E(z) = \sum_{p\in \mathcal{V}} \psi_d(z_p) + \sum_{p\in \mathcal{V}}\sum_{q\in
\mathcal{E}_p}\psi_s(z_p,z_q), \label{eq:jointcrf}
\end{align}
where $z$ is the variable set $z_p = [c_p,m_p]$. $c_p$ denotes selection of the $c_p$th
regional correspondence for pixel $p$ and $m_p$ is the segmentation label. $\psi_d$ and
$\psi_s$ are the unary and pair-wise potentials. $\mathcal{V}$ is the set including all
image pixels and $\mathcal{E}_p$ denotes image pixels for the fully connected CRF.

\vspace{0.1in}\noindent\textbf{Joint Unary Potential $\psi_d(z_p)$~~} The new part in
this potential is to model the correspondence and segmentation interaction prior. It is
defined as
\begin{align}
\psi_d(z_p) = \psi_d^j(c_p,m_p) + \alpha_1\psi_d^c(c_p) + \alpha_2\psi_d^m(m_p),
\label{eq:crfunary}
\end{align}
where $\psi_d^j(c_p,m_p)$ models the joint potential between the $c_p$ and $m_p$ in pixel
$p$. $\psi_d^c(c_p)$ and $\psi_d^m(m_p)$ are the potentials on the correspondence and
segmentation likelihood respectively. $\alpha_1$ and $\alpha_2$ weight the three terms. A
larger $\alpha_1$ emphasizes correspondence more and $\alpha_2$ influences segmentation
likelihood.

We define the joint potential $\psi_d^j(c_p,m_p)$ according to the joint distribution
\begin{align}
\psi_d^j(c_p,m_p) = -\log \big(h(w(c_p),m_p)\big),
\end{align}
where $w(c_p)$ is for the $c_p$th dominant correspondence for pixel $p$. $h(w,m)$ is the
joint distribution between correspondence and segmentation. Since we have initialization
correspondence and segmentation, we estimate $h(w,m)$ by computing the joint histogram.

For the regional correspondence unary potential $\psi_d^c(c_p)$, we define it based on
the matching cost. Motivated by optical flow intensity and gradient constancy, the
potential is defined as
\begin{align}
\psi_d^c(c_p) = 1-\exp\Big({-\mu(I_1,I_2,w(c_p))}/{\sigma_c^2}\Big),
\end{align}
with {\small
$$
\mu(I_1,I_2,w(c_p)) = \|I_{1,p}-I_{2,p+w(c_p)}\|+\|\nabla I_{1,p}-\nabla
I_{2,p+w(c_p)}\|,
$$}
where $\mu$ computes the matching cost and $\nabla$ is the gradient operator. $\|\cdot\|$
computes the $L_1$ distance. $\sigma_c$ is the parameter controlling the matching cost.
We set it to 0.2 in all our experiments.

We model the segmentation unary potential $\psi_d^m$ by the FCN predicted probability,
which is defined as
\begin{align}
\psi_d^m(m_p) = -\log\big(S(m_p)\cdot C(m_p)\big),
\end{align}
where $S(m_p)$ indicates the probability of pixel $p$ taking label $m_p$. We compute the
probability using FCN predicted score after soft-max normalization. Rather than directly
using original FCN model, we fine-turn it with our labeled portraits, which will be
detailed later. $C(m_p)$ is estimated from the foreground and background color model.
With the initial segmentation mask, we fit a Gaussian mixture model (GMM) for color
distributions of foreground and background as $h_b(x)$ and $h_f(x)$, similar to those of
\cite{Liu2009_paintselection}. With the color models, we set $C(m_p) =
(1-m_p)h_b(I_{1,p}) + m_ph_f(I_{1,p})$. In all our experiments, we apply four Gaussian
kernels for the background and six for foreground.

\begin{figure}[t]
\centering
\begin{tabular}{@{\hspace{0.0mm}}c@{\hspace{1mm}}c@{\hspace{1mm}}c@{\hspace{1mm}}c@{\hspace{0mm}}}
\includegraphics[width=0.24\linewidth]{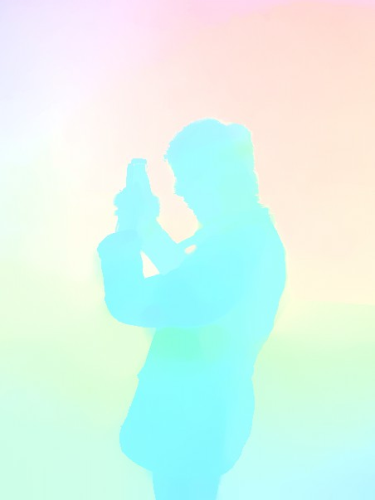}&
\includegraphics[width=0.24\linewidth]{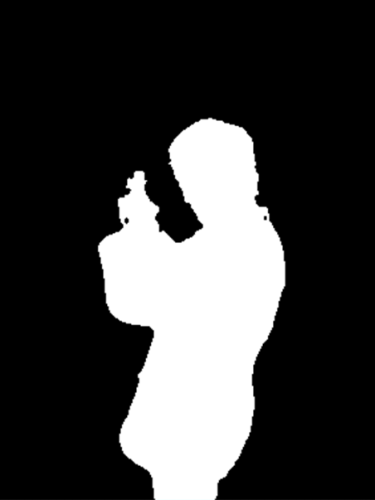}&
\includegraphics[width=0.24\linewidth]{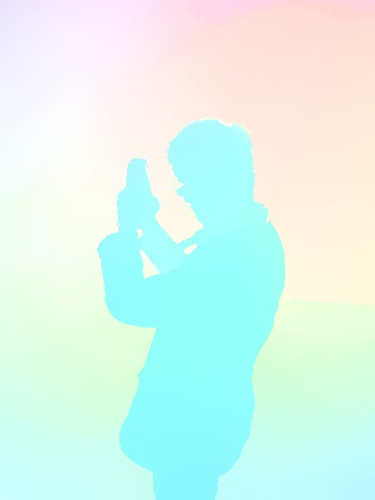}&
\includegraphics[width=0.24\linewidth]{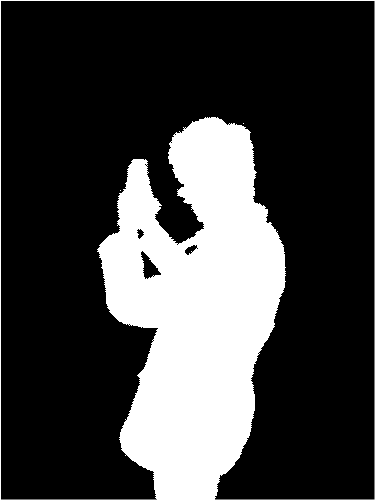}\\
\small(a) Sep. Corr. & \small (b) Sep. Seg. & \small(c) Joint Corr.  & \small(d) Joint Seg.\\
\end{tabular}
\caption{Comparison of separate and joint refinement on correspondence and segmentation
using our CRF model. (a) and (b) are separately refined correspondence and segmentation
results. (c) and (d) are the joint ones. The input is shown in Figure
\ref{fig:textureless}(a). } \label{fig:jointcompare}
\end{figure}

\vspace{0.1in}\noindent\textbf{Joint Pair-Wise Term $\psi_s(z_p,z_q)$~~} The pair-wise
term enforces regional flow selection and segmentation labeling for piece-wise
smoothness. The correspondence and segmentation should have similar smooth property with
close discontinuity in both images. To achieve it, the pair-wise term is formulated with
the following three items.
\begin{align}
\psi_s(z_p,z_q) &= \beta_1 \psi_s^c(c_p,c_q)\psi_s^m(m_p,m_q) + \beta_2 \psi_s^c(c_p,c_q)\nonumber \\
&+ \beta_3 \psi_s^m(m_p,m_q). \label{eq:pairwise}
\end{align}
The first item is joint pair-wise smoothness between $c$ and $m$. The goal is to force
consistency between segmentation and correspondence. The last two items are the
smoothness penalty in regional correspondence and segmentation labels respectively.
$\beta_1$, $\beta_2$ and $\beta_3$ are the parameters. Similar to those of
\cite{Li2004_lazy,Liu2009_paintselection,Krahenbuhl2011}, we define them using the Potts
model with bilateral weights as {\small
\begin{align}
&\psi_s^c(c_p,c_q) = \delta(|c_p\!-\!c_q|)\cdot g(\|p-q\|,\|I_{1,p}\!-\!I_{1,q}\|,\sigma_s,\sigma_r),\nonumber\\
&\psi_s^m(m_p,m_q) = \delta(|m_p\!-\!m_q|)\cdot
g(\|p-q\|,\|I_{1,p}\!-\!I_{1,q}\|,\sigma_s,\sigma_r),
\end{align}
}where $\delta(x)$ is zero when $x$ is zero and is one otherwise.
$g(x,y,\sigma_s,\sigma_r)$ is the bilateral weight function defined as
$\exp({-x^2}/{\sigma_s^2}-y^2/\sigma_r^2)$. The weight enforces neighboring pixels with
similar color to select the same label in correspondence and segmentation space.
$\sigma_s$ and $\sigma_r$ are the spatial and range parameters, which have the same
influence as those in bilateral filter \cite{TomasiM98}.

\subsection{Inference}

The objective function defined in Eq. (\ref{eq:jointcrf}) is an $NP$-hard problem on two
sets of valuables $c$ and $m$. To efficiently infer them, we separate the system into two
sub ones on $c$ and $m$ and alternatively update estimation.
\begin{itemize}\vspace{-0.08in}
  \item Given correspondence $c^t$, we optimize segment $m^t$.\vspace{-0.08in}
  \item With updated segmentation $m^t$, we solve for $c^{t+1}$.\vspace{-0.08in}
\end{itemize}
$t$ indexes iterations. The two sub-problems can be solved efficiently by mean field
approximation \cite{Krahenbuhl2011_fastcrf}. In our experiments, 3-4 iterations are
enough to get satisfying results.

\subsection{Analysis}

\noindent\textbf{Why Joint Form?~~} The proposed joint model for correspondence and
segmentation refinement makes use of correspondence labeling and segmentation. We compare
it with separately processing correspondence and segmentation. In Eq.
(\ref{eq:jointcrf}), the joint model degenerates to independent refinement when omitting
all terms with respect to $c_p$ and $m_p$ respectively. We evaluate these models, and
show results in Figure \ref{fig:jointcompare}. It is noticeable that separately refining
labels performs less well than our current system. Estimation of correspondence and
segmentation can benefit each other via utilizing their mutual information.

\begin{figure}[t]
\centering
\begin{tabular}{@{\hspace{0.0mm}}c@{\hspace{1mm}}c@{\hspace{0mm}}}
\includegraphics[width=0.48\linewidth]{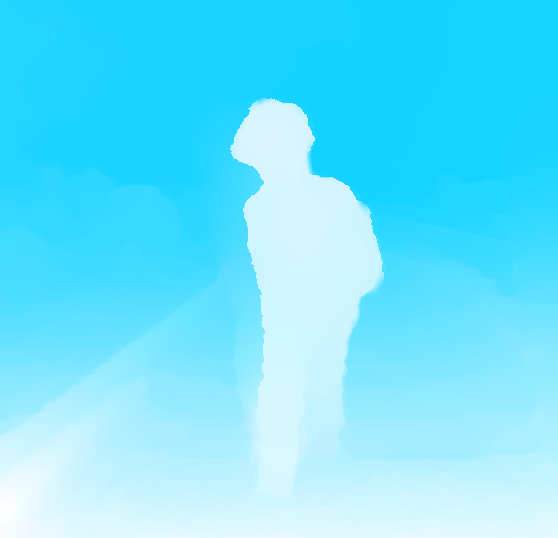}&
\includegraphics[width=0.48\linewidth]{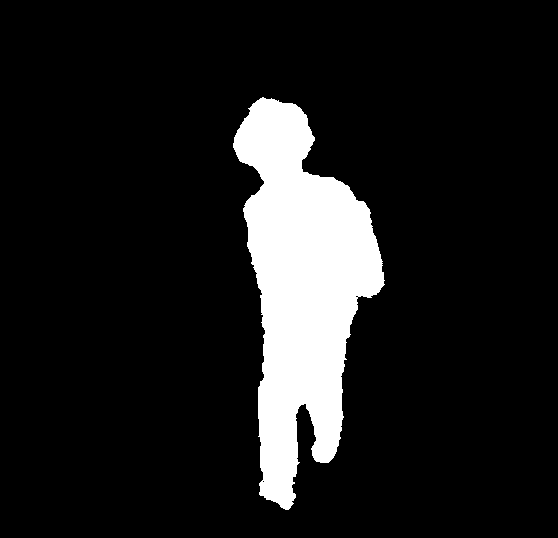}\\
\small(a) MRF Corr. & \small (b) MRF Seg. \\
\includegraphics[width=0.48\linewidth]{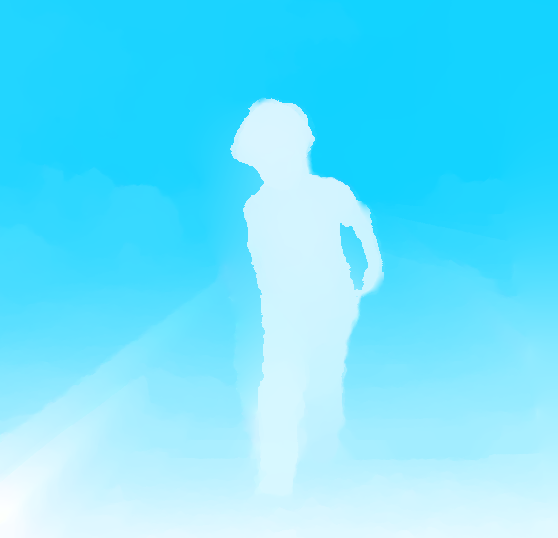}&
\includegraphics[width=0.48\linewidth]{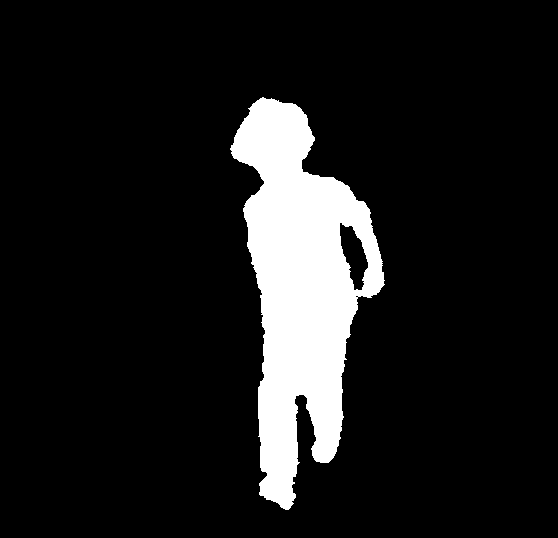}\\
\small(c) CRF Corr. & \small(d) CRF Seg.\\
\end{tabular}
\caption{Comparison of MRF and fully connected CRF. (a) and (b) are the results with the
MRF pair-wise term. (c) and (d) are the results with our fully connected CRF term. The
input image is shown in Figure \ref{fig:teaser}. \vspace{-0.1in}} \label{fig:graphcutcrf}
\end{figure}

\vspace{0.15in}\noindent\textbf{Fully Connected CRF~~} Compared with general MRF, which
uses only 4- or 8-neighbor smoothness terms, the fully connected CRF has the ability to
label a very small region if it is globally distinct. To illustrate it, we show a
comparison in Figure \ref{fig:graphcutcrf}. For the results in (a) and (b), our model
with the MRF term cannot correctly obtain the arm area because the region is very small.
In contrast, our method is based on fully connected CRF and can handle such cases, as
shown in (c) and (d).

\vspace{0.15in}\noindent\textbf{Difference from Previous Approaches~~} Our method
automatically refines semantic segmentation and correspondence estimation. Methods of
\cite{SunSB10,SunSB12,SunWSPB13} applied the layer information to higher quality
correspondence inference. But no semantic object information is applied. Method of
\cite{BaiLKU16} exploited segmentation to help correspondence estimation. However,
segmentation results are not refined in following processing. In addition, approach of
\cite{Sevilla-LaraSJB16} aims to model motion patterns for objects while ours is to
simultaneously and effectively refine human segmentation and dual-lens correspondence.

\section{Evaluation and Experiments}

\begin{figure}[t]
\centering
\begin{tabular}{@{\hspace{0.0mm}}c@{\hspace{1mm}}c@{\hspace{1mm}}c@{\hspace{1mm}}c@{\hspace{0mm}}}
\includegraphics[width=0.24\linewidth]{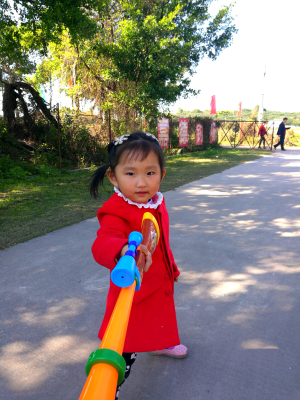}&
\includegraphics[width=0.24\linewidth]{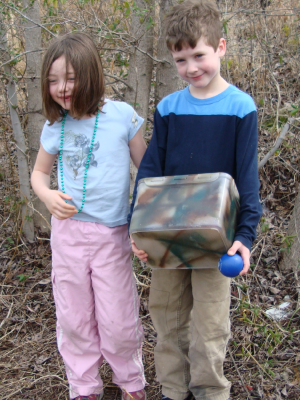}&
\includegraphics[width=0.24\linewidth]{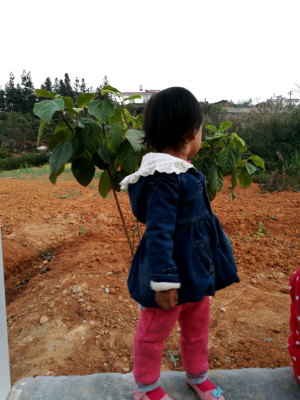}&
\includegraphics[width=0.24\linewidth]{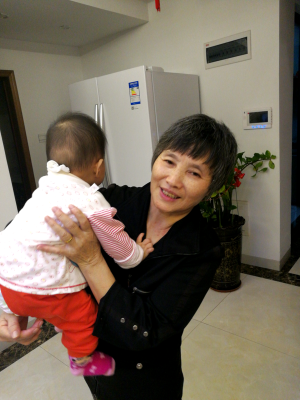}\\
\includegraphics[width=0.24\linewidth]{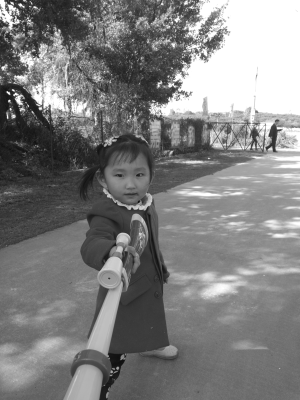}&
\includegraphics[width=0.24\linewidth]{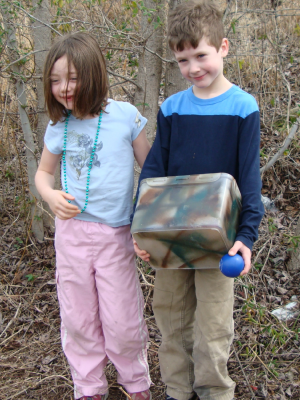}&
\includegraphics[width=0.24\linewidth]{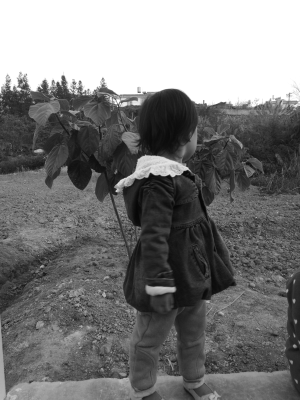}&
\includegraphics[width=0.24\linewidth]{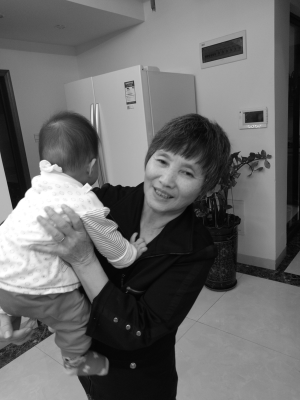}\\
\end{tabular}
\caption{Our dual-lens image examples. The images in the top row are from the left camera
and the bottom ones are from the right camera of Huawei P9.} \label{fig:dataexamples}
\end{figure}

We collected dual-lens portrait images with a Huawei P9 smart phone. We also search the
data from Flickr with key words ``stereo'' and ``3D image''.  A few examples are shown in
Figure \ref{fig:dataexamples}. We select persons with a large variety in terms of age,
gender, clothing, accessory, hair style and head position. Image background is with
diverse structure regarding locations of indoor and outdoor scenes, weather, shadow, etc.
All our captured images are with resolution $3968\times2976$. Between the two captured
images, one is with color and the other is grayscale because of the special camera
setting. We denote the color image as reference and the grayscale one as input. Portrait
areas are cropped and resized to $1200\times800$. $2,000$ portrait image pairs are
collected, which include 1,850 captured ones and 150 downloaded from Flickr.

With the selected dual-lens portrait images, we first label the human body segments in
the color reference image using Photoshop quick selection tool \cite{AdobePhotoshop} and
take them as portrait segment ground truth.

Since it is very difficult to achieve accurate image correspondence, we fuse
different-algorithm results with user interaction. First, we obtain correspondence
results using state-of-the-art optical flow methods MDP \cite{XuJM12}, DeepFlow
\cite{Weinzaepfel2013}, EPPM \cite{BaoYJ14flow}, and LDOF \cite{BroxLOD10}. For each
method, we choose eight groups of parameter values and finally get 32 correspondence maps
for each image pair. Second, we select the best correspondence from the 32 candidates
using the method of \cite{LempitskyRR08}. Third, we label unmatched area with user
interaction and apply flow completion \cite{LiuYT014_steadyflow}. Finally, we take edited
correspondence maps as ground truth for all portrait image pairs. We split the 2,000
pairs into 1,800 pairs for training and 200 for evaluation.

\subsection{Comparison and Evaluation}

In terms of the system structure, we compute initial Horn-Schunck optical flow using the
code of \cite{SunRB10} with default parameters. Fast weighted median filter
\cite{ZhangXJ14wmf} is then applied to smooth it. For semantic segmentation
initialization, we changed the original FCN-8s model to 2 outputs, which are the
background and foreground similar to that of \cite{Shen2016EG}. Then the model is fine-
tuned using our training data based on the original FCN-8s model. The fine-tuning process
can improve segmentation accuracy, to be shown below. For the joint update model, we set
$\alpha_1$ and $\alpha_2$ both to 1.5. $\beta_1$, $\beta_2$ and $\beta_2$ are all set to3
by default. $\sigma_s$ ranges from 10 to 20 and $\sigma_r$ is set around $0.2$. The
running time of our method for a $600\times 800$ image pair is 16.63 seconds on an Intel
Core-i7 CPU PC without any GPU acceleration. In all our experiments, the results are
generated in 3 iterations.

\vspace{0.05in} \noindent\textbf{Evaluation on Our Data~~} With our data, we evaluate the
methods quantitatively in terms of segmentation and correspondence accuracy. We compare
the person segmentation with state-of-the-art methods FCN \cite{Long2014}, DeepLab
\cite{Chen2014_deeplab} and CRFasRNN \cite{Zheng2015_crfrnn} using the author published
model. Besides directly applying the original 20-class object model, we change each model
to 2-class output with portrait and background. These methods are all fine-tuned with our
portrait data. We define these fine-tuned models as ``FCN-portrait'',
``DeepLab-portrait'' and ``CRFasRNN-portrait''.

The results are reported in Table \ref{tab:segres} where we apply the
intersection-over-union (IoU) to measure the segmentation accuracy with respect to ground
truth. The table shows that the three 20-class object segmentation models achieve around
80\% IoU accuracy. By updating the models to 2-class output and further fine-tuning them
by our portrait data, their accuracy is improved by about 4\%. We also test our model
only updating the segmentation, which achieved very limited improvement. Our joint model
presents the best performance, bearing out the effectiveness of jointly refining
correspondence and segmentation.

\begin{table}[t]
\small \centering
\begin{tabular}{|l|c|}
  \hline
  {Methods} & ~~~Mean IoU(\%)~~~\\
  \hline
  \hline
  FCN \cite{Long2014}                   & 79.51 \\
  DeepLab \cite{Chen2014_deeplab}               & 80.09 \\
  CRFasRNN \cite{Zheng2015_crfrnn}              & 80.23 \\
  FCN-portrait            & 83.90 \\
  DeepLab-portrait        & 84.01 \\
  CRFasRNN-portrait~~~~~~~~~~~~~~ & 84.19 \\
  Ours-separate & 84.32 \\
  \hline
  \textbf{Ours}                 & \textbf{88.33} \\
  \hline
\end{tabular}\vspace{0.05in}
\centering \caption{Comparison of segmentation results on our data. ``FCN-portrait",
``DeepLab-portrait'' and ``CRFasRNN-portrait'' denote fine-tuned models using our labeled
image data. ``Ours-separate'' is the model only updating
segmentation.\vspace{-0.1in}}\label{tab:segres}
\end{table}

\begin{table}[t]
\centering \small
\begin{tabular}{|l|c|c|}
  \hline
  Methods & ~~~AEPE~~~ & ~~~AAE~~~\\
  \hline
  \hline
  HS Flow \cite{SunRB10}     &  13.66     & 10.48  \\
  TV-L1 Flow \cite{BroxBPW04}     & 10.01      & 8.52   \\
  LDOF Flow \cite{BroxLOD10}     & 8.32      & 7.81  \\
  MDP Flow \cite{XuJM12}     &  8.23     & 7.96  \\
  EPPM Flow \cite{BaoYJ14flow}~~~~~~~~~~~~~     &  11.74     & 9.05  \\
  DeepFlow \cite{Weinzaepfel2013}     &  7.87     & 6.81  \\
  EpicFlow \cite{Revaud2015_epic}     &  8.11      &7.49   \\
  Ours-separate     &  8.03      &7.45   \\
  \hline
  \textbf{Ours}          &  \textbf{5.29}     & \textbf{5.91}  \\
  \hline
\end{tabular}\vspace{0.05in}
\caption{Comparison of correspondence results on our data. We calculate the average end
point error (AEPE) and average angular error (AAE).}\label{tab:correspondence}
\end{table}

We compare our methods with other dense correspondence estimation approaches, including
Horn-Schunck \cite{SunRB10}, TV-L1 \cite{BroxBPW04}, MDP \cite{XuJM12}, DeepFlow
\cite{Weinzaepfel2013}, LDOF \cite{BroxLOD10}, EpicFlow \cite{Revaud2015_epic}, and EPPM
\cite{BaoYJ14flow}. Evaluation results are given in Table \ref{tab:correspondence}.
Compared with the variational model without feature matching constraints, such as HS and
TV-L1 model, the methods LDOF, MDP, DeepFlow, and EpicFlow achieve better performance. We
also evaluate our model by only refining the correspondence. The result is much improved
over the initial HS flow. Our final joint model yields the best performance among all
matching methods.

\begin{figure*}
\centering
\begin{tabular}{@{\hspace{0.0mm}}c@{\hspace{.5mm}}c@{\hspace{.5mm}}c@{\hspace{.5mm}}c@{\hspace{.5mm}}c@{\hspace{0mm}}}
\includegraphics[width=0.185\linewidth]{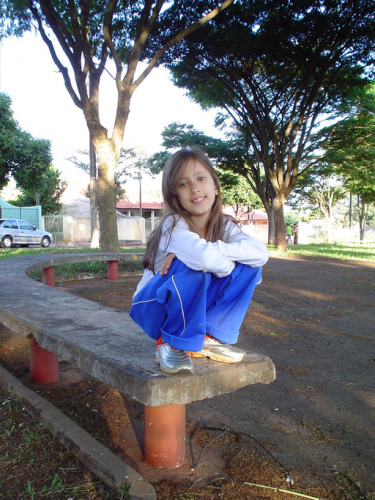}&
\includegraphics[width=0.185\linewidth]{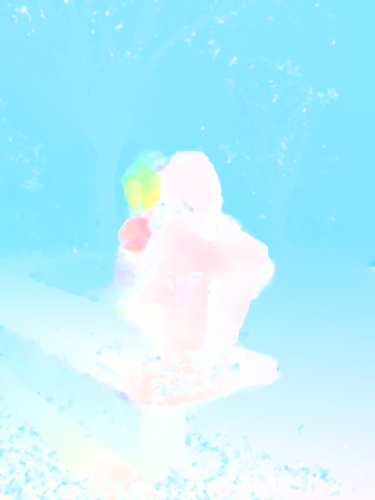}&
\includegraphics[width=0.185\linewidth]{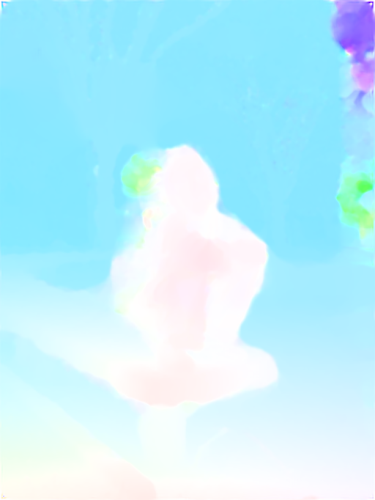}&
\includegraphics[width=0.185\linewidth]{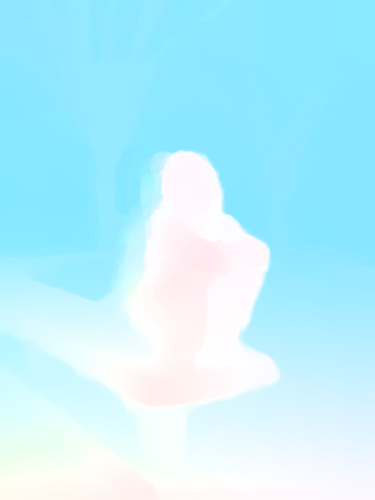}&
\includegraphics[width=0.185\linewidth]{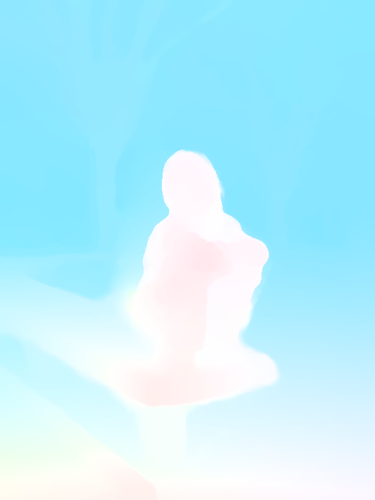}\\
\small (a) Input & \small (b) MDP Flow &\small (c) LDOF Flow &\small (d) DeepFlow &\small (e) Our Corr. \\
\includegraphics[width=0.185\linewidth]{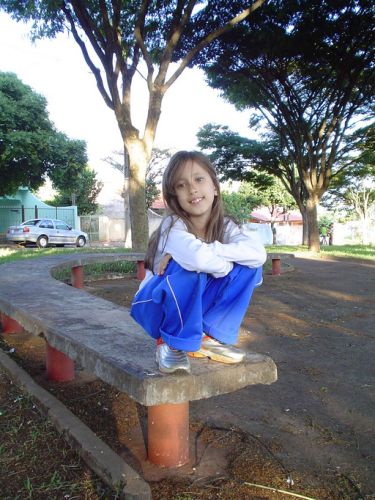}&
\includegraphics[width=0.185\linewidth]{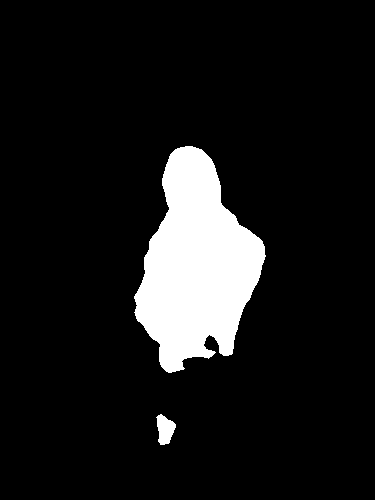}&
\includegraphics[width=0.185\linewidth]{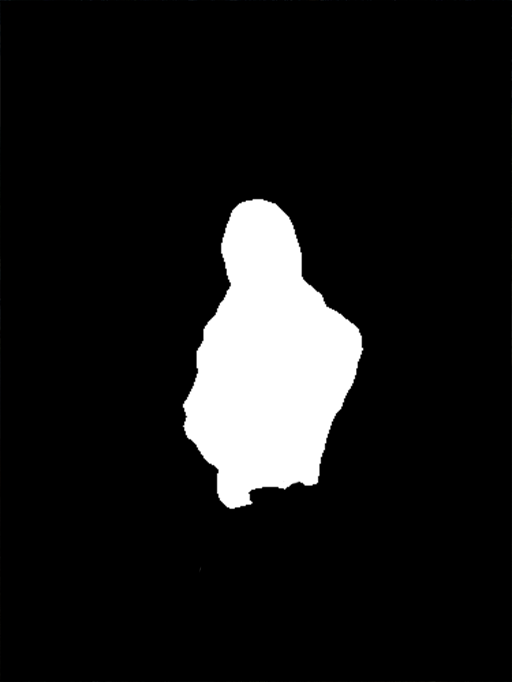}&
\includegraphics[width=0.185\linewidth]{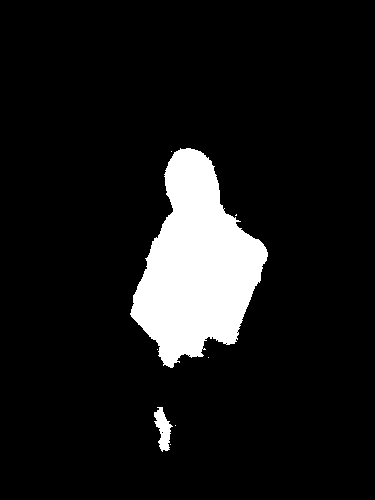}&
\includegraphics[width=0.185\linewidth]{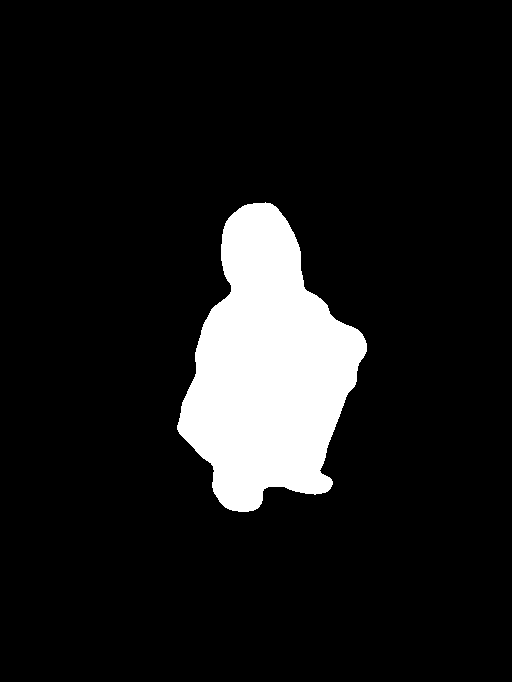}\\
\small (f) Reference & \small (g) FCN &\small (h) FCN-portrait &\small (i) CRFasRNN &\small (j) Our Seg. \\
\includegraphics[width=0.185\linewidth]{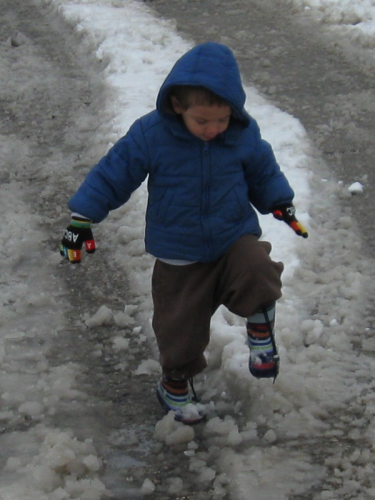}&
\includegraphics[width=0.185\linewidth]{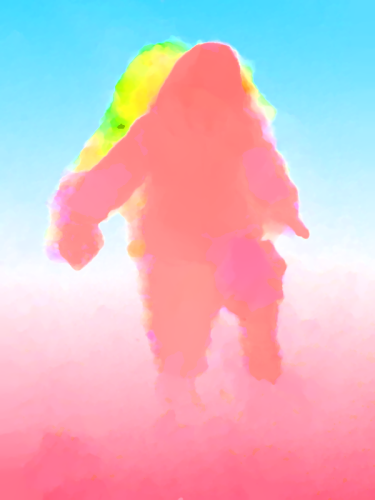}&
\includegraphics[width=0.185\linewidth]{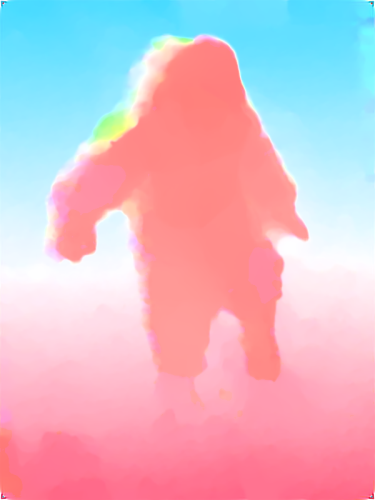}&
\includegraphics[width=0.185\linewidth]{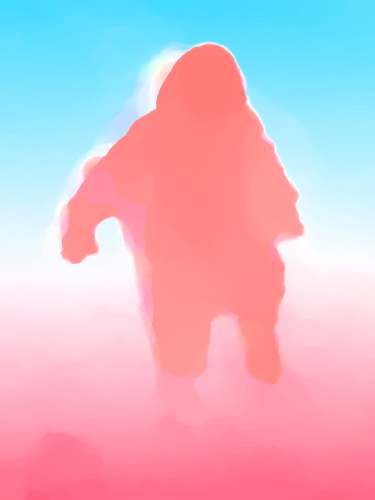}&
\includegraphics[width=0.185\linewidth]{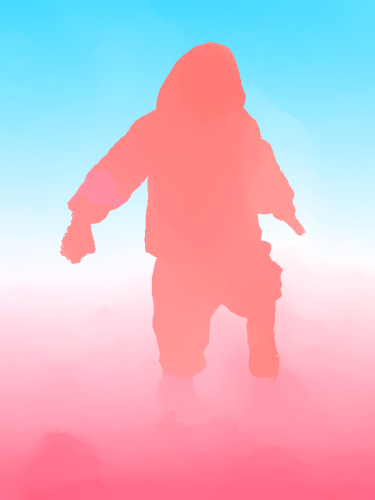}\\
\small (a) Input & \small (b) MDP Flow &\small (c) LDOF Flow &\small (d) DeepFlow &\small (e) Our Corr. \\
\includegraphics[width=0.185\linewidth]{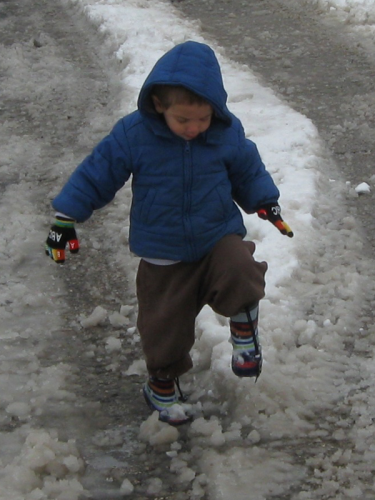}&
\includegraphics[width=0.185\linewidth]{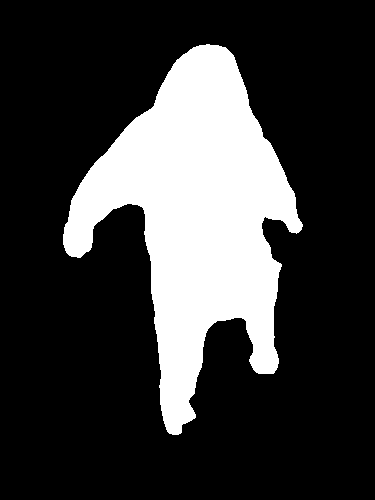}&
\includegraphics[width=0.185\linewidth]{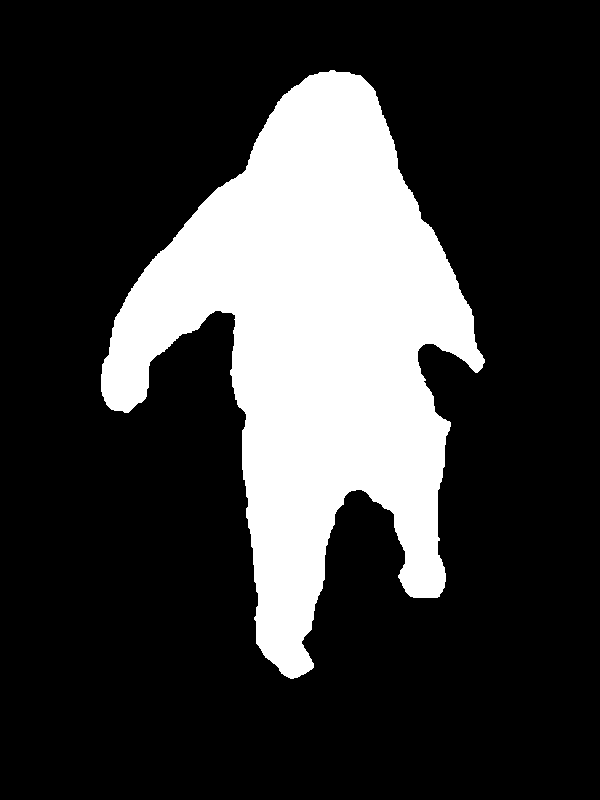}&
\includegraphics[width=0.185\linewidth]{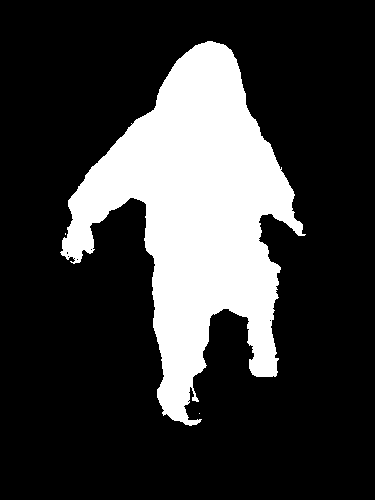}&
\includegraphics[width=0.185\linewidth]{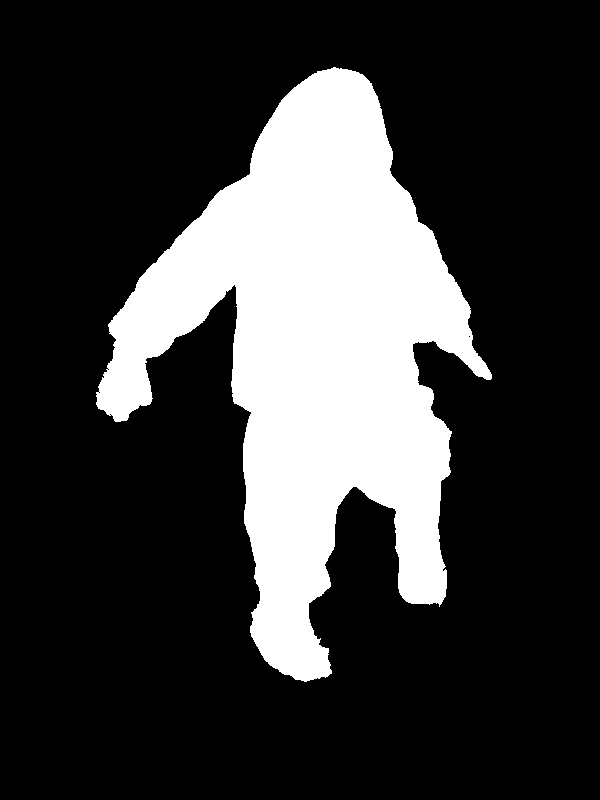}\\
\small (f) Reference & \small (g) FCN &\small (h) FCN-portrait &\small (i) CRFasRNN &\small (j) Our Seg. \\
\end{tabular}
\caption{Visual comparisons of different segmentation and correspondence estimation
methods. (a) and (f) are the input and reference images respectively. (b-e) are the
correspondence estimated by MDP \cite{XuJM12}, LDOF \cite{BroxLOD10} and DeepFlow
\cite{Weinzaepfel2013} respectively. (g-i) are the FCN, FCN-portrait and CRFasCNN
segmentation results respectively. (e) is our correspondence result and (f) is our
segmentation result.\vspace{-0.1in}} \label{fig:visualcmp}
\end{figure*}

\vspace{0.05in} \noindent\textbf{Visual Comparison~~} As shown in Figure
\ref{fig:visualcmp}, we compare our method with previous matching methods MDP
\cite{XuJM12}, LDOF \cite{BroxLOD10}, DeepFlow \cite{Weinzaepfel2013} and semantic
segmentation approaches FCN \cite{Long2014}, FCN-portrait, and CRFasRNN
\cite{Zheng2015_crfrnn}. Our method also notably improves the matching accuracy in human
body boundaries and textureless regions. By utilizing the reliable correspondence
information, decent performance is accomplished for portrait segmentation.

\vspace{0.05in}\noindent\textbf{Regional Correspondence Effectiveness~~} Our regional
correspondence estimation greatly speeds up the labeling process and increases accuracy
by resolving the textureless issue. To verify it, we compare our method with the
non-regional estimation scheme, which is to set $w^i$ into discrete constant maps
covering all possible displacements. We use 500 uniformly sampled values from
$[-50,50]\times[50,50]$ to get all $w^i$s. As reported in Table \ref{tab:whyrc}, with our
regional correspondence, the method is 10 times faster and is also more accurate in terms
of the AEPE measure.

\begin{table}[t]
\small \centering
\begin{tabular}{|l|c|c|}
  \hline
  {Methods} & Accuracy (AEPE) & Running Time (Seconds)\\
  \hline
  \hline
  without RC &6.45 & 186.3\\
  \hline
  with RC  &5.29 & 16.63\\
  \hline
\end{tabular}\vspace{0.05in}
\centering \caption{Performance of our regional correspondence estimation. ``RC'' denotes
the regional correspondence. \vspace{-0.1in}}\label{tab:whyrc}
\end{table}

\section{Conclusion}
We have proposed an effective method for joint correspondence and segmentation estimation
for portrait photos. 
Our method still has the following
limitations. First, our approach may fail when the image contains many persons -- our
training data does not include such cases. Second, the extra low-level imaging problems
such as highlight, heavy noise, and burry could degrade our method for reliable
correspondence and segmentation estimation. Our future work will be to deal with these
issues with more training data and enhanced models.

{\small
\bibliographystyle{ieee}
\bibliography{matching}
}

\end{document}